# Logistic Ensemble Models

Bob Vanderheyden
Department of Statistics and Analytical Sciences
College of Science and Mathematics
Kennesaw State University

Jennifer Priestley
Department of Statistics and Analytical Sciences
College of Science and Mathematics
Kennesaw State University

*Abstract*—**Predictive models that are developed in a regulated industry or a regulated application, like determination of credit worthiness, must be interpretable and "rational" (e.g., meaningful improvements in basic credit behavior must result in improved credit worthiness scores). Machine Learning technologies provide very good performance with minimal analyst intervention, making them well suited to a high volume analytic environment, but the majority are "black box" tools that provide very limited insight or interpretability into key drivers of model performance or predicted model output values. This paper presents a methodology that blends one of the most popular predictive statistical modeling methods for binary classification with a core model enhancement strategy found in machine learning. The resulting prediction methodology provides solid performance, from minimal analyst effort, while providing the interpretability and rationality required in regulated industries, as well as in other environments where interpretation of model parameters is required (e.g. businesses that require interpretation of models, to take action on them).**

## I. Introduction

The current study compares a combination of *k* less effective logistic regression models (predictors/classifiers), called an "ensemble," to a fully developed model. In specifying the fully developed predictor, techniques that are commonly applied to maximize the performance of a single model (e.g., nonlinear transformations) are applied. The combination of *k* predictors are of the form:

$$p = \lambda_1 p_1 + \lambda_2 p_2 + \cdots + \lambda_k p_k \qquad (3)$$

Where $p_i$ is the probability estimate of the ith model (predictor) and $\lambda_i$ is the weight, determined by the optimization used to create the ensemble results.

The fundamental hypothesis for the study is:

A composition of multiple logistic regression classifiers, using no analyst derived attribute transformations or analyst determined attribute selection steps* will perform as well or better than an optimally developed logistic regression model, in the original time period and performance will be more stable (better), than the optimally developed logistic regression model across an extended time period.

\* - no analysis/data manipulation beyond basic cleansing and imputation of missing values

Considerable published research demonstrates the value of using modern machine learning methodologies for predictive tasks like the classification of data records into one of two known classes. As such, machine learning methodologies provide superior performance for tasks like identification of prospective customers/buyers.

Additional research demonstrates that moving beyond single classifiers (models) to develop ensembles of machine learning classifiers provides additional performance benefit in terms of prediction accuracy. Abellán, et al [1], and Zhou, et al, [2] provide excellent examples of this kind of advancement. More specific to this study, Doumpos, et al studied the impact of creating stacked generalization ensembles for seven different machine learning methodologies in attempting to improve performance of credit scoring models [3].

While these studies report enhanced prediction performance for ensembles of models, the algorithms do not provide a capability that allows for the interpretation of model changes due to changes in an underlying data attribute (e.g., the impact on credit score of changes a key credit behavior attributes like credit card balance).

When the predicted event falls under the purview of government regulation, or if the company leadership requires insight into the "key drivers" of a predictive model, machine learning methodologies aren't viable options for predictive model development, since the contribution of specific data attributes and impact of changes in an attribute cannot be determined empirically.

In the United States, there are laws and regulations focused on ensuring and monitoring fair access to credit [4]. These laws exist to protect specific classes of individuals and small businesses that have experienced significant and debilitating discrimination in the past. Financial institutions must develop lending programs that are designed to avoid any violations of those laws and regulations [4]. For consumer and some small business lending, a predetermined sufficient set of relevant factors for loan approval and even pricing may be embedded in a custom credit scoring. In many cases, credit bureau scores are key determining factors in credit decisions [4]. Differences in lending rates and pricing for protected and unprotected classes of consumers and small businesses must be examined to ensure that systematic lending decisions aren't biased [4]. A key investigative tool is the detailed examination of predictive models, that form the basis of credit scores that are included in these decisions (e.g., credit scores). [4]

In an effort to incorporate relatively recent machine learning advances in prediction, large corporations, including credit bureaus, are investing in adaptations of machine learning methodologies that include underlying algorithmic processes to ensure that model behavior conforms to expected behavior (e.g., the impact of changes in credit behavior on credit score). Companies that are industry leaders in the development of applied analytics methodologies have the resources to develop custom machine learning methodologies, like Equifax's soon-to-be patented NeuroDecision risk modeling technology [5]. Equifax's solution includes the reporting of "reason codes" that can be used to inform prospective customers of the specific elements in their credit history that are impacting the credit decision.

Many companies aren't able to invest in these kinds of customer machine learning algorithms. These companies are essentially relegated to either purchasing services, or using analytic methods like logistic regression that have been industry standards for decades, but tend to not perform as well as machine learning methodologies.

Logistic regression is a well studied, successful statistical method for predicting a binary event (i.e., an event with two possible outcomes). The methodology estimates the log-odds of the event based on linear regressors, using the functional form:

$$\ln\left(\frac{p}{1-p}\right) = \boldsymbol{\beta x} \quad (1)$$

Where $p$ is the estimated probability of an event, $\boldsymbol{x}$ is a vector of available predictive data elements and $\boldsymbol{\beta}$ is a vector of weights determined by the logistic regression algorithm.

For model estimation purposes, this functional form is transformed to express the binary event in terms of the logit function:

$$y = \frac{1}{1 + e^{-\boldsymbol{\beta x}}} \quad (2)$$

This same logit function is used as the activation function (represented by nodes in the model below), in many neural nets, deep nets and convolutional neural nets. The simplest case of a neural net, is the multilayer perceptron (MLP).

In the figure 1, the nodes (circles) represent the logistic regression (or other) activation function and the arrows represent data fed into the node. In "nesting" these logistic regression models, the MLP is a complex nonlinear model. The weights or coefficients of the models cannot be interpreted. This lack of interpretive ability makes neural nets inappropriate for credit worthiness models.

Fig 1: Multi-Layer Perceptron Structure

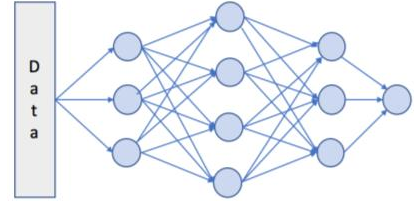

One critical concern for any predictive model is its general applicability. Developing a "better" performing model that doesn't generalize to new data provides limited value for the business or agency that needs to make decisions based on the application of the model. In many situations, models perform well within time, but when applied to a future time period, model performance suffers and performance further deteriorates as the time from initial model development increases [6].

In machine learning, generalizability of modeling techniques that tend to suffer from issues related to over fitting can be mitigated by employing one or more ensemble methodologies. The process combines results of multiple lower performing models to provide a high performing solution that generalizes better than a single model. One example of this technique is the development of random forests which combine the results of multiple decision trees [7].

## II. Literature Review

The concept of developing "ensembles" of models has been around for more than 30 years. "Stacking" is one of the primary forms of ensemble model creation. This methodology uses the same training sample for all of the classifiers. Different models are created using different modeling methodologies or using the same methodology, but with different predictive elements [8].

In 1992, David Wopert published foundational research examining the stacking of neural net models, to boost generalizability of predictive models [8]. The process of using a classifier to combine a set of classifiers into an ensemble is a common practice in machine learning called "stacking" [9].

For scenarios where multiple predictors are available, choosing the "best" single predictor may not generate the best result [8]. Instead, Wolpert performs a heuristic analysis of the viability of using the results from multiple predictors, that he positions as a "generalization" of predictors.

Wolpert starts with the most fundamental assessment of generalization of predictive models – the validation sample – and then extends his analysis to generalization of a predictor to other populations/samples. Wolpert not only provides a heuristic foundation supporting the use of stacked models, but includes two key experiments that demonstrate the applicability and value [8].

Wolpert's work is a general discussion of and strong support for combing predictors, but he doesn't arrive at a single "best practice" for combining predictors. In later work, Wolpert and Macready develop the famous "No Free Lunch (NFL) Theorems," which conclude that a best practice that covers all problems doesn't exist [10]. The NFL Theorems were a motivating reason that Doumpos and Zopounidis investigated the benefits of creating ensembles of different types of classifiers [3].

In their work, Doumpos and Zopoundis, use the same classifiers to create the ensembles, in the second stage. For each classifier type (logistic regression, neural net, etc), they test all available classifiers, including all parameter scenarios (e.g., number of nodes and layers in a neural net), in an effort to determine the optimal ensemble. They then select the best of the set of stacked classifiers to represent the initial classifier. As a benchmark, they compare the results to the best initial classifier, based on cross validation [3].

Due to multicollinearity concerns, Doumpos and Zopoundis perform PCA (Principal Components Analysis) on the initial set of predictions, prior to creating the ensemble. They keep all components that explain a minimum of 10% of the combined variance [3]. It should be noted that utilizing PCA, in a predicitive model, for a regulated application, isn't acceptable, since a clear understanding of precisely what is happening within the model, relative to data inputs isn't possible (i.e., violating the interpretability requirement).

In 1992, Leo Brieman published "Stacked Regression" which examines simple linear combinations of predictors of the same type. These combinations were of the form [11]:

$$v(x) = \sum_k \alpha_k v_k(x) \qquad (4)$$

Unrestricted, $\alpha_k$'s could lead to a combined solution that isn't as effective as one or more of the available predictors [11]. Restricting the parameters to be non-negative and sum to one results in what Brieman calls an "interpolating" predictor that performs at least as well as the best single predictor.

To estimate the optimal $\alpha_k$, Brieman minimizes the quadratic cost function, subject to two conditions.

$$\sum_n (y_n - \sum_k \alpha_k v_k(x))^2 \qquad (5a)$$

$$\sum \alpha_k = 1 \qquad (5b)$$

$$\alpha_k \geq 0 \qquad (5c)$$

Brieman also notes that after optimization is complete, a relatively small number of predictors have a non-zero weight [11]. This sum of squared error cost function for the linear combination of predictors will be used to optimize the ensemble that's produced in this study.

In his work, Brieman examined stacking multiple predictors from five different classes of predictors (a single class within each of five separate analyses) [11]. This study focuses on stacking a variation of Brieman's "Subset Regressions" which combines predictors that utilize different sets of predictive elements, based on a random sampling of available data elements, to derive an optimal predictor.

Instead of using Ordinary Least Squares or Ridge Regression, per Brieman's work, this study examines Subset Logistic Regression. In addition, this study creates the Subset Logistic Regression based on stratified time samples within the provided data file structure (i.e., samples from individual quarters) and compares the ensemble results to an optimal single model that utilizes data elements across four quarters.

III. DEVELOPING LOGISTIC REGRESSION CLASSIFIERS

For the purposes of this study, credit data for 11.8 million prospective business customers are analyzed. The quarterly data, with over 300 data points per quarter, was provided by a major U.S. based credit bureau. The data span 9 years from 2006 to 2014. This data set offers an opportunity to not only assess performance of a predictive binary classification model versus the proposed solution within a specified time frame, but also allows for the assessment of model performance over an extended period of time.

One of the credit default related attributes was selected to produce credit default binary classes, for accounts in 2007. Businesses with similar default behavior, in the prior year (2006), are highly likely to have an account in default in the target year (2007). These repeat offenders would provide an artificial "lift" to model performance, so they are removed from the analysis. Another way of looking at this is that it would be irrational for a lender to extend credit to a prospective customer who is already in default on other loans, so these accounts are removed from the analysis.

As is standard procedure for virtually every parametric model development effort, basic data cleansing and missing value imputation was completed before the model development processes started.

For the benchmark or "Base Model," variable clustering using SAS Proc Varclus [12] was performed on the 300 independent data elements, in each of four quarters. The results were use to reduce the set of candidate variables [13]. The results from the analysis of each quarter were then combined and an additional variable clustering, followed by selection of candidate variables was completed.

Continuous independent variables were binned in two ways: 1) using SAS Proc Rank [14] and 2) user defined cut points [15]. Odds and log odds, of the dependent variable, were calculated for the each of the newly derived bins, creating four new sets of variables [15]. Each transformed variable is analyzed to determine if adjacent bins can be combined (collapsed) into a single bin [15].

The binning and creation of nonlinear transformed data elements multiplied the number of available independent values roughly six fold (not all variables were binned). Increasing the number of independent variables by a factor of six resulted in significant redundancy of information, so another variable clustering procedure was conducted to reduce the number of variables to 53. From this point, a logistic regression model was estimated using SAS Proc Logistic [16]. After a series of standard assessments to remove trivial contributors (900K records can result in statistically significant predictive elements that have almost zero actual contribution to the model), the final model contained 22 independent variables that contributed to prediction.

The full model development effort for the Base Model required the equivalent of two weeks of full time work and required an analyst who possessed significant training and experience in order to achieve a good performance level. Both of these factors cannot be understated since in contrast, the proposed solution not only performs better than the Base Model, but required significantly shorter development time and minimal experience for the analyst.

### IV. LOGISTIC ENSEMBLE METHODOLOGY

The fundamental premise of model (classifier) ensembles is the aggregation of the results of multiple sub-optimal models with an application of a "winning strategy" to result in the final prediction. For classification problems, like the binary classification examined in this study, strategies range from simple voting (classify a record on all models then count the yes vs no results and classify accordingly) to more sophisticated strategies like using the model results as inputs to an additional binary classifier (i.e, stacking) – similar to the functionality of a simple neural net [9].

For the purposes of this study, an optimal linear combination of the predictions for the set of logistic regression models is used. The strategy of creating an ensemble that uses a linear combination of plays a critical role in providing interpretability and rationality as required for regulated industries.

As stated earlier, the primary goal of this research is to determine if an ensemble of classifiers based on minimal analyst intervention can produce results that are on par with the more involved model development process used to develop the Base Model. To achieve this goal, for each quarter, data element samples were drawn from the 300+ raw data points that were available after data cleansing and missing value imputation. This stratification of data into quarters provides and additional performance hurdle for the proposed solution, since data for different quarters won't be included in any of the individual models. No variable reduction procedures were utilized.

For each quarter, 40 samples of the data elements were drawn. For each, a random sample of 25% of the available data elements, was create. The logistic procedure in SAS (though any competing product could be used), with simple backwards elimination was used to develop a total 160 models (four models for each of 40 quarters). Each model attempts to predict the defined default event, in the next calendar year. No additional model/variable assessments were used to improve individual model performance or to reduce model generalization concerns like multi-collinearity.

After the 160 models were estimated SAS's Proc LP procedure [17] was used to determine the optimal linear combination of models that maximize prediction. A quadratic program, using the least squares cost function, similar to the one found in Ordinary Least Squares (OLS) regression, was used. The specific problem is:

$$min \sum_{over\ T} (y - \lambda_1 p_1 - \lambda_2 p_2 - \cdots - \lambda_{160} p_{160})^2 \quad (6a)$$

$$st\ \lambda_1 + \lambda_2 + \cdots + \lambda_{159} + \lambda_{160} = 1 \quad (6b)$$

$$\lambda_i \geq 0 \quad (6c)$$

where *T* is the training set, $p_i$ is the logistic regression probability estimate for the *ith* model and $\lambda_i$ is the weight for the ith model.

Note that quadratic programming attempts to identify *an* optimal solution (there may be more than one) and uses different algorithms than OLS. OLS attempts to identify *the* optimal soluion using a closed form of the algorithm that includes $(X'X)^{-1}$ to solve the optimization problem. If a subset of the columns of X are correlated, $(X'X)^{-1}$ does not exist. If columns are highly correlated, then estimation of $(X'X)^{-1}$ may cause problems and inflate weights in the model.

In quadratic programming, even if 2 or more of the suboptimal predictors are perfectly correlated, the algorithms arrive at an optimal solution (though there would be multiple optimal solutions). Therefore, we don't need to be concerned with multicolinarity, per Doumpos and Zopoundis, in their research [3].

To simplify this step, a closed form of the squared error function was calculated by expanding the quadratic function. The resulting quadratic program is then solved. There are alternate ways to specify the model, that would likely be easier to develop an algorithm, if full automation of the process is desired.

For 160 models, the closed form of the objective function requires over 13,000 coefficient combinations ($_{160}C_2$ $\lambda_i\lambda_j$ terms where $i \neq j$ + 160 $\lambda_j^2$ terms + 160 $\lambda_j$ terms ). Solving the quadratic program identified 22 models that contribute to the optimal combination of values. For others, the $\lambda'_i$s are 0. This result is consistent with Breiman's findings, discussed earlier. For each record in the validation set and for each subsequent year, the 22 models with non-zero weights ($\lambda'_i$s

) are scored and the weights are used to generate the linear combination of model scores (i.e., the ensemble score).

Development of the ensemble model required two days to complete. The 160 models were built using a SAS Macro, to allow the process to run while other activities were completed. The quadratic program required roughly 2 hours to set up and identified the optimal solution within 5 minutes of submission. This result represents a significant time saving for an experienced data scientist. In addition, since minimal intervention is required, it is possible to automate the entire procedure, other than data cleansing, missing value imputation and identification of dependent and independent variables. Clearly, if the same data cleansing and missing value imputation processes are always used, even those activities could be automated. Time required for respecifying the models at a future date or to use the same database to create a model that predicts a different event could be as small as an hour or less.

## V. FINDINGS

### A. Base Model Performance

The Base Model has good performance. The 22 predictive data elements produced a model that has a percent concordance of 85.2 and a KS statistic of 54:

Fig 6: Base Model Performance

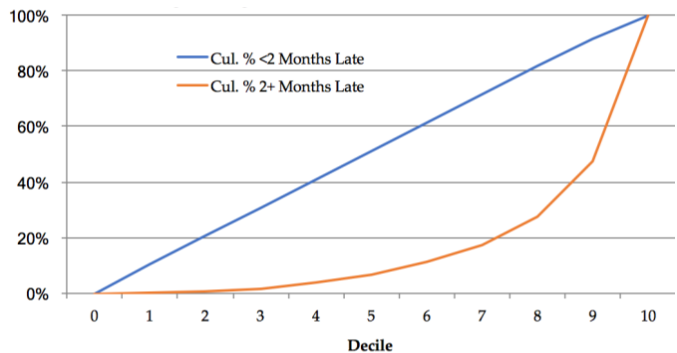

Fig 6. Base Model percent of "bad accounts" by model score decile in the model ordered validation sample vs the percent of "good accounts"

Somewhat surprisingly, the Base Model had very impressive performance when applied to the additional years. These additional years of data represent not only a lag forward in time (one type of generalization challenge), but also due to an increasing number of available records (each quarter adds over 100K new accounts to the database), for scoring, the additional years of data represented an assessment of generalization on new prospective accounts. Performance not only didn't degrade but had a modest increasing trend, in terms of the KS statistic:

Fig 7: Base Model Performance Over Time

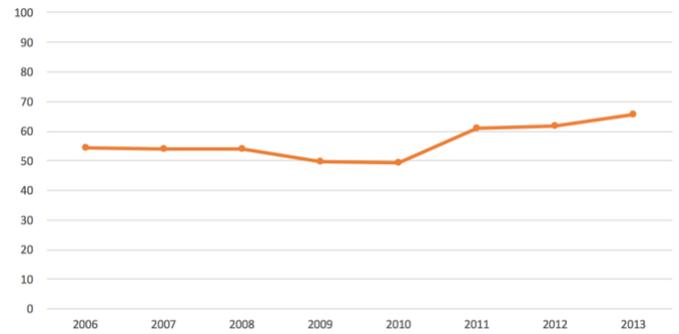

Fig 7. Base Model KS statistic for application of the model to independent samples, over time.

### B. Ensemble Model Performance

The ensemble of quarterly logistic regression models also had good performance. While percent concordance isn't available, the KS statistic can be calculated for the ensemble. When applied to the 2006 validation set, the Ensemble Model outperformed the Base Model and had a KS statistic of almost 58 (a 7% improvement):

Fig 8: Ensemble Model Performance

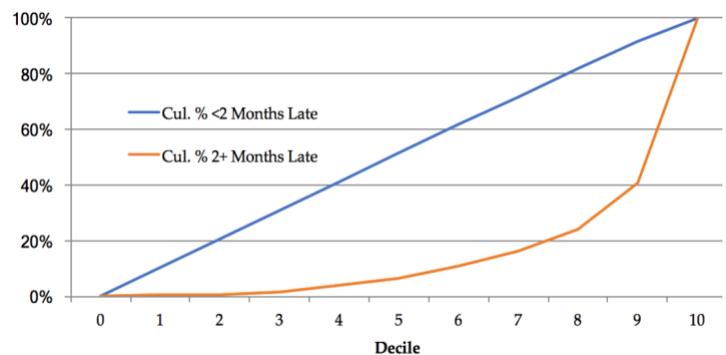

Fig 8. Ensemble Model percent of "bad accounts" by decile in the model ordered validation sample vs the percent of "good accounts"

While higher performance in the same time period is good, the most important assessment is the application over time. As can be seen in the chart below, the Ensemble Model continued to outperform the Base Model in each of the available years of data.

Fig 9: Ensemble Model Performance

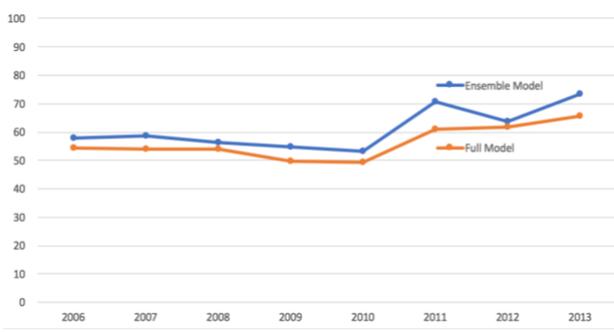

Fig 9. Base Model KS statistic for application of the model to independent samples, over time.

*C. Coefficient Interpetation*

For a single logistic regression model, an analyst interprets a coefficient in terms of change in odds, for a unit change in single data value, for record of data (***d***). In its simplest terms, if the ith element of ***d*** is incremented by 1,

$$\ln\left(\frac{p}{1-p}\right) = \boldsymbol{\beta}(\boldsymbol{d} + \boldsymbol{e_i}) \qquad (7)$$

$$\frac{p}{1-p} = e^{\boldsymbol{\beta d}} e^{\beta_i} \qquad (8)$$

So change the ith element of ***d***$_i$ by 1 unit and the odds will change by a multiple of $e^{\beta_i}$.

Given the nature of the ensemble, calculating change in odds isn't viable. Instead, analysts can lean on multivariate calculus and derive the rate of change with respect to a given data element. For n predictors in the optimal ensemble:

$$\frac{\partial p}{\partial x_i} = \lambda_1 \frac{\partial p_1}{\partial x_i} + \lambda_2 \frac{\partial p_2}{\partial x_i} + \cdots + \lambda_n \frac{\partial p_n}{\partial x_i} \qquad (9)$$

$$= \lambda_1 \frac{\beta_{1i} e^{-\beta_1 x}}{(1+e^{-\beta_1 x})^2} + \cdots + \lambda_n \frac{\beta_{ni} e^{-\beta_n x}}{(1+e^{-\beta_n x})^2} \qquad (10)$$

$$= \lambda_1 \beta_{1i} p_1(1-p_1) + \cdots + \lambda_n \beta_{ni}\, p_n(1-p_n) \qquad (11)$$

From here, we arrive at a close form for estimating the change in p as a funciton of changing $i$th data element of a sample data record:

$$\Delta p = \left(\lambda_1 \beta_{1i} p_1(1-p_1) + \cdots + \lambda_n \beta_{ni}\, p_n(1-p_n)\right)\Delta x_i \quad (12)$$

While this calculation is more complicated than what is required for a single logistic regression model, the resulting insight is a direct change in model score or credit score (if the model scores are converted to credit score metrics) for a change in identified parameter for a specific business behavioral profile.

For general reporting purposes (e.g., reports to regulators), the mean or median value for the data elements could be used.

## VI. CONCLUSIONS

The current study represents a successful model performance enhancement by blending a powerful machine learning enhancement process with a tested traditional statistical predictive methodology. The Base Model performed very well, but given the 3 separate variable clustering processes that were used to reduce the number of data elements, may have reduced the available variance (i.e, "explanatory power" of the predictive data elements) by 28%, since roughly 15% of variance is removed in each of the two initial variable clustering steps. A significant amount of this lost information was likely accounted for by the nonlinear transformations. Secondarily, the development of the base model required several weeks of work by an experienced data scientist.

The greatly reduced time required to develop the Ensemble Model as well as the very low analyst expertise/intervention adds to its appeal. In fact, the Ensemble Model development process could easily be developed in to a full automated process that would greatly benefit organizations that need to build large numbers of classifiers (e.g. separate classifiers for 100 products or for 100 different countries).